## Title

# Nationality encoding in language model hidden states: Probing culturally differentiated representations in persona-conditioned academic text

## Author details

Paul Jackson[1], Ruizhe Li[2], and Elspeth Edelstein[3]

[1] Language Centre, School of Language, Literature, Music and Visual Culture, University of Aberdeen, Aberdeen AB24 3FX, United Kingdom.

[2] School of Natural and Computing Sciences, University of Aberdeen, Aberdeen AB24 3FX, United Kingdom.

[3] School of Language, Literature, Music and Visual Culture, University of Aberdeen, Aberdeen AB24 3FX, United Kingdom.

## Abstract

Large language models are increasingly used as writing tools and pedagogical resources in English for Academic Purposes, yet it remains unclear whether they encode culturally differentiated representations when generating academic text. This study investigates whether

Gemma 3 4b it, a 4 billion parameter instruction tuned transformer, encodes nationality discriminative information in its hidden states when generating research article introductions conditioned by British and Chinese academic personas. A corpus of 270 texts was generated from 45 prompt templates crossed with six persona conditions in a $2 \times 3$ design. Logistic regression probes were trained on hidden state activations across all 35 layers, with shuffled label baselines, a surface text skyline classifier, and cross family generalisation tests used as methodological controls. Probe selected token positions were annotated for structural, lexical, and stance features using the Stanza NLP pipeline, and a sentence level baseline tested whether any observed differences were visible in the full generated surface text.

The nationality probe achieved a cross validated accuracy of 0.968 at Layer 18 with perfect held out classification. Nationality encoding followed a non monotonic trajectory across layers, with structural effects strongest in the middle to upper part of the network and lexical domain effects peaking earlier. At probe selected token positions, British associated patterns showed higher rates of postmodification, hedging, boosting, passive voice, and evaluative or process oriented vocabulary, while Chinese associated patterns showed higher rates of premodification, nominal predicates, and sociocultural or internationalisation vocabulary. Critically, the sentence level baseline found no significant nationality differences in the full generated surface text. The model therefore encodes culturally differentiated rhetorical patterns in its hidden representations and at specific high signal token positions, but these distinctions are not statistically visible in the output as a whole.

These findings show that nationality encoding in language model representations is structured, layer specific, and linguistically characterisable, but that it does not surface robustly in the generated text. This pattern is consistent with culturally hollow writing, in which output displays interactional fluency without cultural grounding. The results partly align with and partly diverge from cross cultural differences documented in the applied linguistics literature, suggesting that the model has internalised distributional regularities from its training data that correspond to some, but not all, attested rhetorical tendencies. The study extends probing methodology to a sociolinguistic attribute not previously examined in this way and identifies EAP and language pedagogy as a domain in which hidden cultural encoding has direct practical consequences for multilingual students and educators who cannot detect those assumptions through surface reading alone.

## 1.Introduction

Large language models are increasingly used in academic writing as drafting tools, feedback sources, and pedagogical models for multilingual students in higher education. Their outputs are often treated as examples of competent academic prose. Research on generative AI in language education reports conditional benefits for drafting, vocabulary development, and engagement, but recent work on academic discourse also shows that AI-generated writing can differ from expert writing in move structure, stance management, and rhetorical range (Barrot, 2023; Kong and Liu, 2024; Mo and Crosthwaite, 2025). Cultural-alignment research further suggests that these outputs are not culturally neutral, but reflect patterned value orientations shaped by training data and prompting conditions, with closer alignment to English-speaking and Western European contexts than to many other national settings (AlKhamissi et al., 2024; Tao et al., 2024).

These findings raise a question for interpretability research. Language models are known to encode syntactic, morphological, and semantic information in hidden representations, and probing classifiers are a standard method for testing what is recoverable from those representations across layers (Belinkov, 2022; Tenney et al., 2019). However, cultural and nationality-related issues in language models have more often been studied at the level of outputs than through probing of hidden states, and probing-based work on sociolinguistic or cultural attributes such as nationality has received limited attention. It therefore remains unclear whether the cultural differentiation observed in model outputs reflects structured internal encoding or emerges through mechanisms not accessible to linear probing.

This question has practical significance in English for Academic Purposes and language pedagogy within engineering higher education, where academic writing norms are transmitted to multilingual students moving between educational systems. If a model encodes nationality-linked rhetorical patterns when generating text about writing development, feedback, or assessment, those patterns may shape what students encounter as standard academic practice.

The contrastive rhetoric literature provides an empirical benchmark for evaluating such patterns. Research has documented cross-cultural differences in hedging, boosting, modification, and discourse organisation in English and Chinese academic writing (Hinkel, 2002; Hu and Cao, 2011; Loi and Evans, 2010), while Hyland's work shows why stance, hedging, and interactional positioning matter more broadly in academic discourse (Hyland, 1998, 2004, 2005b). These findings provide directional expectations that can be tested against the features of probe-selected tokens. At the same time, the model is not a human writer but a statistical system shaped by the distributional properties of its training data. Alignment with documented cross-cultural tendencies would suggest that comparable regularities are present in the training data; divergence would indicate that the model has learned nationality-linked associations that do not map closely onto established human patterns.

The study uses Gemma-3-4b-it, a 4-billion-parameter instruction-tuned transformer, to generate 270 research article introductions conditioned by British and Chinese nationality personas crossed with instructional medium and academic role. Hidden-state activations are captured across all 35 layers, and logistic regression probes are used to predict nationality from centroid and token-level representations. Controls include shuffled-label baselines, a surface-text skyline classifier, and cross-family generalisation tests. Probe-selected tokens are annotated with the Stanza NLP pipeline for structural, lexical, and stance features, and a sentence-level baseline tests whether any nationality differences are visible in the full generated surface text or are concentrated at positions where nationality is most strongly encoded.

The study addresses four research questions:

RQ1. Does Gemma-3-4b-it encode nationality-discriminative information in its hidden representations when generating persona-conditioned academic text?

RQ2. At which layers is nationality encoding strongest, and how does this signal vary across network depth?

RQ3. What structural, lexical, and stance features characterise the tokens or positions at which nationality encoding is strongest?

RQ4. Are nationality-linked differences visible in the full generated surface text, or are they concentrated mainly at probe-selected positions?

## 2. Literature review

### *2.1 Academic writing, interculturality, and cross-cultural variation*

Academic writing is not a neutral or purely technical skill (Swales, 1990, 2004; Hyland, 2000, 2005a, 2005b). It is shaped by communicative purpose, disciplinary discourse, interpersonal positioning, and institutional power (Lea and Street, 1998, 2006; Lillis and Curry, 2010; Canagarajah, 2002). This matters for the present study because any account of AI-generated academic writing must consider rhetorical organisation and stance, not surface fluency alone.

The cross-cultural dimension of academic writing is well established. Hinkel (2002) documents systematic variation in hedging, syntax, and rhetorical patterning across L1 backgrounds, while Hyland (1998) shows that hedging in research writing is tied to the social practices of academic communities rather than to uncertainty alone. In direct comparisons of English and Chinese academic writing, Loi and Evans (2010) find differences in research article introduction organisation, and Hu and Cao (2011) show that abstracts in Chinese-medium applied linguistics journals contain fewer hedges and more boosters than those in English-medium journals. At the level of phraseology and discourse organisation, Chen and Baker (2010), Pan, Reppen and Biber (2016), and Liu and Furneaux (2014) identify systematic differences in lexical bundles, modification patterns, and text development. Landwehr (2025) further shows, in English scientific writing, that premodification and postmodification are distinct strategies of information packaging rather than superficial stylistic variants, with noun phrase complexity interacting systematically with modification type. Taken together, these studies provide directional expectations against which model-internal differences in modification and rhetorical organisation can be interpreted.

From a pedagogical perspective, intercultural communicative competence is central to language education rather than an optional addition (Byram, 1997, 2021; Deardorff, 2006; Kramsch, 1993; Bennett, 1993; Spitzberg and Changnon, 2009). These frameworks treat intercultural development as involving reflexivity, judgement, and relational awareness rather than formal correctness alone. Formal fluency in a second language, or in machine-generated text, therefore cannot be treated as equivalent to intercultural adequacy (Byram, 1997, 2021; Deardorff, 2006; Kramsch, 1993).

### *2.2 Generative AI, academic discourse, and culturally hollow writing*

Research on generative AI in language education has expanded rapidly since the release of ChatGPT, but most empirical work has focused on writing and short-term higher-education settings (Zhu and Wang, 2025; Li et al., 2024; Li et al., 2025; Lo et al., 2024). Reported benefits are conditional rather than automatic: ChatGPT and related systems can support drafting, idea generation, vocabulary development, and engagement, and learners' positive perceptions of AI use are associated with reduced writing anxiety (Yılmaz and Üstünel, 2025), but broader learning effects depend on task design, checking procedures, collaboration, and teacher mediation rather than on access to the tool alone (Barrot, 2023; Warschauer et al., 2023; Yan, 2023).

Recent work in The Journal of English for Academic Purposes (JEAP) narrows this picture to academic discourse. Kong and Liu (2024) show that AI-generated abstracts differ from scholar-written abstracts in move structure. Mo and Crosthwaite (2025) show that large language models can reproduce stance and engagement features, but with a narrower rhetorical range than human writers. Hyland (2026) argues that generative AI reproduces the formal markers of academic interaction without communicative intent, producing interactional fluency without relational grounding. These findings indicate that coherent academic output may still differ from expert writing in ways that matter for EAP, especially in the management of stance, hedging, boosting, and interpersonal positioning (Hyland, 2005a, 2005b).

A recurrent problem in this literature is that AI-generated writing may be fluent without being culturally or pragmatically grounded. Output can appear coherent, structured, and lexically appropriate while remaining limited in cultural depth, pragmatic fit, and social grounding, a condition that can be defined as culturally hollow writing (Byram, 1997, 2021; Deardorff, 2006; Kramsch, 1993). Ahmad et al. (2024) provide a concrete example: ChatGPT could produce grammatically plausible responses about Hausa culture and emotion while failing to capture culturally grounded judgements. The issue is therefore not factual error alone, but the gap between fluent text and situated cultural meaning (Ahmad et al., 2024).

### *2.3 Cultural alignment, bias, and inequality in language models*

Cultural-alignment studies make this problem more precise at the level of model behaviour. Tao et al. (2024) show that major GPT models align more closely with English-speaking and Protestant European countries than with many other national contexts. AlKhamissi et al. (2024) show that alignment depends on prompt language, prompting, and the linguistic composition of model pretraining. Model output is therefore not culturally neutral, but reflects patterned value orientations shaped by data and prompting conditions (Tao et al., 2024; AlKhamissi et al., 2024).

This issue sits within a broader problem of representational inequality. Joshi et al. (2020) show that most of the world's languages remain outside the main focus of NLP research, and later multilingual evaluation work shows that this pattern continues in the LLM era (Lai et al., 2023). In educational use, this matters because model output may be treated as authoritative. Where that happens, representational bias becomes part of how rhetorical norms, cultural assumptions, and legitimate forms of knowledge are reproduced (Bender et al., 2021; Noble, 2018; Benjamin, 2019). In the context of academic discourse, dominant Anglophone conventions may marginalise other rhetorical traditions and epistemologies (Canagarajah, 2002; Lillis and Curry, 2010).

### *2.4 Probing hidden representations in language models*

The interpretability literature provides methods for examining what models encode internally rather than only evaluating their outputs. Probing classifiers are a prominent method in this work: a classifier is trained to predict a property from hidden-state representations, and successful prediction is treated as evidence that information relevant to that property is recoverable from those representations (Belinkov, 2022). Belinkov also stresses that high probing accuracy does not establish causal use during generation, and that baselines, control tasks, and upper bounds are needed to distinguish genuine encoding from classifier capacity or geometric artefacts (Belinkov, 2022). Hewitt and Liang (2019), Pimentel et al. (2020), and Voita and Titov (2020) extend this discussion by addressing selectivity, information gain, and probe complexity.

Work that probes representations across layers further shows that transformer representations are hierarchically organised. Lower layers tend to capture more local morphosyntactic information, while higher layers support progressively higher-level linguistic tasks (Tenney et al., 2019). This matters here because different types of nationality-linked signal, if present, may peak at different depths. At the same time, prior probing work has focused mainly on syntactic, morphological, and semantic properties, while cultural or sociolinguistic attributes such as nationality have received much less attention. The present study enters at this point by asking whether nationality, as a persona-level attribute, is encoded in hidden representations when a language model generates academic discourse.

### *2.5 Critical AI literacy and the present study*

Recent work on AI literacy addresses the pedagogical implications of these issues directly. Long and Magerko (2020) define AI literacy as understanding, evaluating, and using AI responsibly. Pérez-Paredes et al. (2025) extends this to applied linguistics by arguing that critical AI literacy must include attention to ethics, language, ideology, and power. Tseng and Warschauer (2023) argue that students need to learn not only how to prompt AI tools, but how to verify and integrate outputs critically. Yan (2023) supports this empirically by showing that the value of ChatGPT-generated feedback depends on how it is processed, discussed, and checked rather than on access to the tool alone.

Taken together, the literature identifies a gap between the surface fluency of AI-generated academic writing and its cultural, rhetorical, and epistemic grounding. Intercultural competence theory explains why fluent text is not necessarily adequate (Byram, 1997, 2021; Deardorff, 2006; Kramsch, 1993). JEAP and related work shows that AI-generated academic discourse can be rhetorically narrower and culturally hollow (Kong and Liu, 2024; Mo and Crosthwaite, 2025; Hyland, 2026; Ahmad et al., 2024). Cultural-alignment studies show that model outputs carry patterned value skews (Tao et al., 2024; AlKhamissi et al., 2024). The probing literature provides methods for testing whether such differences are also structured in hidden representations (Belinkov, 2022; Tenney et al., 2019). What remains underexplored is whether nationality-discriminative information is encoded internally when models generate academic discourse, and if so, what linguistic features carry that signal. This question is especially important in EAP and language pedagogy within engineering higher education, where students may rely on fluent but culturally hollow output as a model of standard academic practice.

## 3. Method

### *3.1 Corpus generation*

The experimental corpus was generated using Gemma-3-4b-it (Google, 2025), a 4-billion-parameter instruction-tuned transformer with 35 layers and hidden dimensionality 2,560. The model was accessed through the Hugging Face Transformers library and run on a single NVIDIA T4 GPU in Google Colab. Greedy decoding was used throughout (do_sample = False), with a minimum output length of 100 tokens and a maximum of 300, ensuring deterministic generation for each prompt.

Forty-five prompt templates were constructed across three families: base, alt, and theory. The base and alt families each requested research-article-style introduction paragraphs of approximately 200 words on EAP, language pedagogy, and academic practice in engineering higher education. The theory family requested longer introductions of approximately 300 words on critical and theoretical topics intersecting engineering with ethics, social theory, and political economy. This domain was selected because the study investigates nationality encoding in the academic and pedagogical discourse through which writing norms are transmitted to multilingual students, rather than in narrowly technical engineering prose.

Templates were paired with six persona clauses derived from a 2 × 3 design crossing nationality (British, Chinese) with cohort type. The cohort variable combined instructional medium and role into three levels per nationality: EMI postdoc, CMI postdoc, and student. Each of the 45 templates was combined with all six personas, yielding 270 unique prompts and 270 generated texts, balanced across nationalities (135 British, 135 Chinese). Text length ranged from 149 to 261 words ($M = 210$). For each generation, hidden-state activations were captured at all 35 layers with output_hidden_states = True. Activations were extracted for generated tokens only, excluding prompt tokens, and saved as compressed NumPy arrays. Although some persona–topic pairings were more ecologically plausible than others, the balanced design ensured that any such mismatch affected both nationalities symmetrically.

### *3.2 Probing architecture*

Two probe types were used. For the example-level probe, token representations for each text were averaged at each layer to form a single 2,560-dimensional centroid vector. A logistic regression classifier with L2 regularisation (C = 1.0, solver = lbfgs, max_iter = 4,000), preceded by z-score standardisation, was trained to predict nationality from these centroid representations. Layer selection used stratified k-fold cross-validation on 216 training examples (80%), with the remaining 54 examples (20%) reserved as a held-out test set evaluated once at the selected layer. The same architecture was also used for medium, role, and six-way cohort classification.

For the token-level probe, the same classifier architecture was applied to individual token representations at each layer. This retained positional granularity and enabled each token to be assigned a signed decision score indicating its nationality-discriminative strength and direction. These scores were then used to identify the tokens carrying the strongest signal.

### *3.3 Token extraction and window construction*

High-signal tokens were extracted from six sampled layers: 2, 10, 18, 24, 30, and 33. These layers span the full depth of the network while sampling more densely in the mid-to-upper region where preliminary probing indicated the nationality signal was strongest. Thresholds were calibrated at Layer 24 and then applied uniformly across all six layers to ensure comparability: the top 2.5% of absolute single-token scores defined the single-token threshold, and the top 5% of mean absolute decision scores within sliding five-token windows defined the window threshold.

For the five-token-window dataset, a sliding window of five consecutive tokens was passed across each generated text at each sampled layer. Windows whose mean absolute score exceeded the threshold were retained, and the central token served as the focus anchor for annotation. Where multiple overlapping windows exceeded threshold, the highest-scoring window was retained for each focus position. This produced 8,015 observations, of which 6,961 remained after quality filtering. For the single-token dataset, individual tokens exceeding threshold were extracted, excluding those already serving as focus anchors in the window dataset. This yielded 2,921 observations, of which 2,353 remained after filtering. The same fixed 80/20 train–test split was used throughout all probing, control, and annotation analyses.

### *3.4 Methodological controls*

Following Belinkov's evaluative framework, three controls were implemented to test whether probe performance reflected genuine representational encoding rather than classifier artefacts (Belinkov, 2022). First, a shuffled-label baseline was computed at the best nationality layer by training the same classifier on 100 random permutations of the nationality labels. Second, selectivity was calculated as the difference between real and shuffled accuracy, following Hewitt and Liang (2019). Third, a surface-text skyline classifier was trained on TF-IDF unigram and bigram features from the generated texts using the same split and evaluation procedure. A hidden-state probe outperforming this skyline indicates access to information not recoverable from lexical distributions alone.

A further cross-family generalisation test assessed whether the nationality signal generalised across prompt content. Probes were trained on one template family and evaluated on each of the other two, producing six train→test comparisons (base→alt, base→theory, alt→base, alt→theory, theory→alt, theory→base), each with balanced nationality labels.

### *3.5 Linguistic annotation*

The probe-selected tokens were annotated using Stanza, which provides tokenisation, POS tagging, lemmatisation, and dependency parsing within the Universal Dependencies framework (Qi et al., 2020). Each token was classified on four structural dimensions: phrase type, modifier structure, clause slot, and predicate type. Stance markers were coded as binary flags for hedging, boosting, modal presence, and passive voice. For the window dataset, a five-token local context was retained around each focus token to support structural interpretation. The single-token and five-token-window datasets were annotated with the same schema so that their outputs could be compared directly.

The Stanza pipeline underwent a veracity audit that identified nine recurrent error classes. Amendments corrected the major systematic parsing problems, including coordination-member propagation, misclassified nominal and adjectival roots, and participial or adnominal structures wrongly labelled as pre-modifiers. Observations flagged as artefactual, low-confidence, or requiring manual adjudication were excluded from statistical analysis. These filters reduced the five-token-window dataset from 8,015 to 6,961 observations and the single-token dataset from 2,921 to 2,353. Because later results test whether the identified nationality signal is visible in the surface text as a whole, a separate sentence-level baseline analysis was also conducted on the full generated texts. Sentence roots were annotated with the same structural schema, yielding 1,900 sentence observations, of which 1,734 passed quality filters. Sentences were additionally classified by opening, middle, and closing position so that stance rates could be compared by position.

### *3.6 Statistical procedures*

For categorical structural variables, nationality differences were tested using chi-square tests of independence, with Cramér's V reported as the effect size; Fisher's exact test was used where expected counts were too small. Continuous decision scores were compared with Mann–Whitney U tests and rank-biserial correlation. Bonferroni correction was applied within related test families. Stance markers were tested with chi-square and odds ratios, again with Bonferroni correction across the four markers.

Five directional contrastive-rhetoric hypotheses were tested: Chinese > pre-modification, British > post-modification, Chinese > nominal predicates, British > adverbial-slot tokens, and British > hedge markers. These predictions were derived from prior work on cross-cultural academic writing (Hinkel, 2002; Hyland, 2004). Additional comparisons were conducted to separate nationality effects from medium and role confounds by testing nationality within EMI-only and CMI-only subsets, and medium within British-only and Chinese-only subsets.

Domain-level lexical analysis used log-odds ratios to identify tokens distinctive to each nationality. Tokens were assigned to six semantic domains: technical, theoretical, sociocultural, pedagogical, research methods, and general. Domain assignment combined manually curated lexicons with Coxhead's (2000) Academic Word List, Ward's (2009) engineering word list, and researcher domain knowledge. Layer-by-layer analyses then tracked chi-square statistics, Cramér's V, and decision-score gaps across the six sampled layers in order to compare structural and lexical trajectories across network depth.

### *3.7 Methodological limitations*

Three limitations are most relevant. First, probing demonstrates recoverability rather than causal use, so the present design cannot show that the identified representations are active in shaping generation (Belinkov, 2022). No intervention method such as activation patching or amnesic counterfactual probing was applied (Elazar et al., 2021; Meng et al., 2022). Second, probe complexity was restricted to a single linear logistic-regression architecture with fixed regularisation, so the study cannot distinguish between compact linear encoding and information recoverable only through more complex decoding (Pimentel et al., 2020; Voita and Titov, 2020). Third, the Stanza pipeline, although audited and amended, inevitably introduces residual classification noise.

### *3.8 Software and reproducibility*

All experiments were conducted in Python on Google Colab with a single NVIDIA T4 GPU. Text generation, activation capture, and probing used NumPy, SciPy, and scikit-learn; linguistic annotation used Stanza with CUDA acceleration; and metadata, activations, and probe outputs were saved locally and mirrored to Google Drive for persistence. The run reported here is run-20260325-110501.

## 4. Results

### *4.1 Corpus descriptives*

The generation procedure produced 270 texts, balanced across nationalities and evenly distributed across the six cohorts. Text length ranged from 149 to 261 words (M = 210), and the British and Chinese subcorpora did not differ significantly in sentence count ($p = .104$). The probe-selected datasets were much larger than the sentence-level baseline because they sampled high-signal positions across six layers rather than full texts only. Cohort frequencies were uneven in the five-token-window data, reflecting differential probe signal strength rather than sampling imbalance; this issue is addressed later through the six-cohort control analyses.

**Table 1. Corpus and analysis datasets**

| Dataset | N before filtering | N after filtering | British | Chinese | Notes |
| --- | --- | --- | --- | --- | --- |
| Generated texts | 270 | 270 | 135 | 135 | 45 per cohort |

| Dataset | N before filtering | N after filtering | British | Chinese | Notes |
|---|---|---|---|---|---|
| Five-token windows | 8,015 | 6,961 | 3,369 | 3,592 | Layers 2, 10, 18, 24, 30, 33 |
| Single tokens | 2,921 | 2,353 | 1,202 | 1,151 | Excludes window anchors |
| Sentence-level baseline | 1,900 | 1,734 | 851 | 883 | Full surface text |

Five-token-window counts by layer were: Layer 2, 1,251; Layer 10, 1,411; Layer 18, 1,105; Layer 24, 1,066; Layer 30, 1,080; Layer 33, 1,048. Cohort counts were: Chinese_Student 2,177; British_Student 1,675; British_EMI_Postdoc 1,525; Chinese_CMI_Postdoc 1,078; Chinese_EMI_Postdoc 338; British_CMI_Postdoc 168.

### *4.2 Probe accuracy and controls*

The strongest evidence for nationality encoding came from the centroid probe at Layer 18. Medium and role also peaked at Layer 18, but with lower accuracy than nationality. All three controls supported the interpretation that the nationality probe was detecting genuine representational structure rather than exploiting label geometry or surface lexical cues. The shuffled-label baseline stayed near chance, selectivity was high, the hidden-state probe far outperformed the surface-text skyline, and cross-family transfer was strong in five of the six train→test combinations. Only base→theory dropped sharply, suggesting that theory prompts activate somewhat different content pathways.

**Table 2. Probe performance and controls**

| Measure | Nationality | Medium | Role |
|---|---|---|---|
| Best layer | 18 | 18 | 18 |
| Cross-validated accuracy | 0.968 ± 0.031 | 0.884 ± 0.044 | 0.940 ± 0.034 |
| Held-out test accuracy | 1.000 | 0.870 | 0.963 |
| Mean shuffled accuracy | 0.499 | 0.504 | 0.575 |
| Selectivity | 0.469 | 0.380 | 0.365 |
| Surface-text skyline, CV | 0.371 ± 0.066 | 0.417 ± 0.004 | 0.676 ± 0.003 |
| Skyline, held-out test | 0.463 | — | — |

Cross-family held-out accuracies ranged from 0.589 to 0.933, with five of six pairings above 0.844.

*4.3 Structural and stance differences at probe-selected positions*

The five-token-window data yielded the clearest structural nationality effects. Phrase type, modifier structure, and UPOS all differed significantly by nationality, while clause slot did not. Chinese-generated high-signal tokens were more often associated with noun phrases and pre-modification; British-generated tokens were more often associated with adjectival or adverbial environments and post-modification. In the stance analysis, British-generated tokens showed higher rates of hedging, boosting, and passive voice, while modal use did not differ significantly.

The single-token data showed a different profile. Pre- and post-modification were not significant, but Chinese-generated tokens showed more nominal predicates and British-generated tokens more adverbial-slot occupancy. Hedge, booster, and passive differences were not significant in this dataset. The contrast between datasets is important: the five-token window captures modifier-rich local contexts, while isolated tokens more often reflect predicate and clause-level positions.

**Table 3. Key nationality contrasts in probe-selected data**

| Comparison | Five-token window | Single token |
|---|---|---|
| Chinese > pre-modifier | Yes: 0.616 vs 0.548, p < .0001, OR = 0.756 | No: 0.414 vs 0.408, p = .801 |
| British > post-modifier | Yes: 0.083 vs 0.062, p = .0008, OR = 1.368 | No: 0.091 vs 0.083, p = .559 |
| Chinese > nominal predicate | No, predicted direction only, p = .704 | Yes: 0.241 vs 0.189, p = .003, OR = 1.361 |
| British > adverbial slot | No | Yes: 0.187 vs 0.142, p = .003, OR = 0.716 |
| British > hedge | Yes: 3.4% vs 2.3%, p = .011, Bonferroni p = .043, OR = 1.463 | No, p = 1.000 |
| British > booster | Yes: 3.1% vs 2.1%, p = .007, Bonferroni p = .028, OR = 1.523 | No, p = .690 |
| British > passive | Yes: 5.2% vs 3.9%, p = .011, Bonferroni p = .044, OR = 1.351 | No, p = .563 |

For the full five-token-window structural tests, phrase type was significant ($\chi^2 = 27.87$, $p < .0001$, $V = 0.063$), modifier structure was significant ($\chi^2 = 46.82$, $p < .0001$, $V = 0.082$), and UPOS was significant ($\chi^2 = 79.93$, $p < .0001$, $V = 0.107$), while clause slot was not ($\chi^2 = 6.05$, $p = .109$, $V = 0.029$).

*4.4 Six-cohort confound control*

The confound checks showed that the nationality signal was not reducible to medium or role. Medium effects within British cohorts and within Chinese cohorts were not significant, and role did not predict phrase type or modifier structure. Nationality within EMI was also non-significant. The only strong subset effect appeared in the CMI-only data, where nationality differences reached significance, but this was driven largely by the small British_CMI_Postdoc subgroup. That pattern warrants caution, but it does not overturn the broader result that medium and role do not account for the overall nationality effect.

**Table 4. Confound-control summary**

| Comparison | Result |
| --- | --- |
| Nationality within EMI | Non-significant: $\chi^2 = 8.98$, $p = .110$, $V = 0.069$ |
| Nationality within CMI | Significant: $\chi^2 = 32.29$, $p < .0001$, $V = 0.161$ |
| Medium within British | Non-significant: $\chi^2 = 9.50$, $p = .091$, $V = 0.075$ |
| Medium within Chinese | Non-significant: $\chi^2 = 5.85$, $p = .322$, $V = 0.064$ |
| Role → phrase type | Non-significant: $\chi^2 = 2.46$, $p = .783$, $V = 0.019$ |
| Role → modifier structure | Non-significant: $\chi^2 = 4.17$, $p = .760$, $V = 0.024$ |

The CMI-only effect should be interpreted cautiously because the minimum expected count was very low and the British_CMI_Postdoc subgroup was small.

### *4.5 Sentence-level baseline*

The sentence-level baseline produced a near-complete null result. No structural variable differed significantly by nationality after correction, and no stance marker showed a reliable nationality effect at the level of full generated sentences. Sentence position affected stance rates, but those effects did not interact with nationality. This contrast with the probe-selected analyses is methodologically central: the nationality differences are detectable at positions where the hidden states encode them most strongly, but not in the generated surface text as a whole.

**Table 5. Sentence-level baseline summary**

| Measure | Result |
| --- | --- |
| Phrase type | $\chi^2 = 0.53$, $p = .768$, $V = 0.017$ |
| UPOS | $\chi^2 = 3.34$, $p = .502$, $V = 0.044$ |
| Predicate type | $\chi^2 = 0.12$, $p = .942$, $V = 0.008$ |
| Hedge rate | British 6.6%, Chinese 8.3%, $p = .275$ |

| Measure | Result |
| --- | --- |
| Booster rate | British 7.4%, Chinese 7.4%, p = .619 |
| Modal rate | British 5.5%, Chinese 5.7%, p = .501 |
| Passive rate | British 15.4%, Chinese 14.1%, p = .815 |

This null result confirms that the probing pipeline is not simply rediscovering broad corpus imbalance. It is isolating positions that are selectively informative for nationality encoding.

### *4.6 Layer trajectory and domain vocabulary*

Nationality-linked differentiation followed a non-monotonic trajectory across layers. In the five-token-window data, the British–Chinese decision-score gap peaked at Layer 10 and again at Layer 18, with smaller gaps at the remaining sampled layers. UPOS differentiation was strongest at Layers 10 and 18, while modifier-structure effects peaked at Layers 2 and 30. This indicates that the nationality signal is not steadily increasing toward the output layer, but is concentrated in specific parts of the network.

The domain-vocabulary analysis showed a clear lexical split. British-distinctive tokens clustered in evaluative and process-oriented discourse, while Chinese-distinctive tokens clustered in sociocultural and internationalisation vocabulary. This effect was significant in the five-token-window data but not in the smaller single-token dataset. Layer-by-layer domain testing showed that the lexical-domain signal peaked earlier than the main structural nationality signal, especially at Layer 10. This suggests that lexical-domain differentiation may be encoded at a lower level of abstraction than the later structural and stance contrasts.

**Table 6. Layer trajectory and domain summary**

| Result | Key values |
|---|---|
| Decision-score gap peak | Layer 10 = 0.290; Layer 18 = 0.200 |
| Highest UPOS Cramér's V | Layer 10 = 0.216; Layer 18 = 0.167 |
| Highest modifier-structure Cramér's V | Layer 2 = 0.143; Layer 30 = 0.133 |
| British domain signal, five-token data | OR = 1.894, $\chi^2 = 39.49$, $p < .0001$ |
| British domain signal, single-token data | OR = 1.426, $\chi^2 = 2.15$, $p = .143$ |
| Strongest layer for domain contrast | Layer 10: 9.6% vs 3.7%, $\chi^2 = 19.13$, $p < .0001$, V = 0.116 |
| Also significant | Layer 24: 9.7% vs 4.9%, $\chi^2 = 8.49$, $p = .004$, V = 0.089 |

Illustrative British-distinctive tokens included stem, reported, recognised, considerable, supervisory, and uncertainty. Illustrative Chinese-distinctive tokens included globalization, international, specialized, oral, instruction, and consensus.

## Discussion

### *5.1 Nationality encoding in hidden states*

The results show that Gemma 3 4b it encodes nationality discriminative information in its hidden representations when generating persona conditioned academic text. This conclusion does not rest on probe accuracy alone. The nationality probe achieved its strongest performance at Layer 18, while the shuffled label baseline remained near chance, the selectivity score was high, the hidden state probe exceeded the surface text skyline, and cross family transfer was strong in five of the six train to test comparisons. Taken together, these findings indicate that the nationality signal is not reducible to prompt wording or lexical overlap, but is present as a structured property of the model's internal representations. In this respect, the study extends probing from its established use on syntactic and semantic properties to a sociolinguistic attribute, while remaining consistent with Belinkov's (2022) caution that probing demonstrates recoverability rather than causal use.

The layer trajectory adds a further dimension to this result. Nationality encoding does not increase steadily toward the final stages of decoding. Instead, it follows a non monotonic pattern, with stronger effects in the middle to upper part of the network and weaker separability closer to output realisation. This is consistent with the general principle established by Tenney et al. (2019) that transformer representations are hierarchically organised and that different kinds of linguistic information become prominent at different depths, though Tenney et al. describe a broadly sequential pipeline rather than a non-monotonic one. In the present study, nationality linked information becomes highly separable before later layers integrate it into the wider decoding process. The significance of this pattern is methodological as well as theoretical. The model forms differentiated internal representations that are only partly visible in the final generated text.

### *5.2 Structural and stance differences*

The structural findings indicate that nationality is encoded through different patterns of information packaging rather than through isolated lexical cues alone. In the five token window data, Chinese high signal positions occurred more often as noun phrase premodifiers, whereas British high signal positions showed higher rates of postmodification. The contrast is therefore one between greater pre head loading and greater post head elaboration. In the single token data, Chinese high signal positions were more likely to occur in nominal predicates, whereas British high signal positions were more likely to occupy adverbial slots. The two datasets reveal different parts of the same broader pattern. The five token window captures modifier rich local contexts, while isolated single tokens more often expose predicate and clause level roles. Nationality is therefore encoded not through one grammatical marker, but through multiple structural tendencies visible at different levels of syntactic granularity.

These findings connect closely to the literature reviewed in Section 2.1. Hinkel (2002) documents systematic cross cultural variation in syntactic and rhetorical patterning. Loi and Evans (2010) show that English and Chinese research article introductions differ in rhetorical organisation, and Hu and Cao (2011) identify differences in hedging and boosting between English medium and Chinese medium journal abstracts. At the level of phraseology and discourse development, Chen and Baker (2010), Pan, Reppen and Biber (2016), and Liu and Furneaux (2014) identify differences in lexical bundling, modification, and text development. Landwehr (2025) further shows, in English scientific writing, that premodification and postmodification are distinct strategies of information packaging rather than superficial stylistic variants. The present study does not reproduce any one of these findings in a one to one way. It does, however, show that the model has internalised a stylised contrast between denser contextual loading and greater post head elaboration, and that this contrast is tied to the nationality conditions encoded in its hidden states.

The stance findings reinforce this interpretation, but they also complicate it. British high signal positions showed higher rates of hedging, boosting, and passive voice in the five token window data. Hyland (1998) argues that hedging in research writing is tied to the social practices of academic communities, and Hyland (2005a, 2005b) shows that academic discourse manages interpersonal meaning through hedges, boosters, attitude markers, and engagement resources.

The hedge result therefore fits reasonably well with the broader applied linguistics literature. The booster result is different. Hu and Cao (2011) found stronger boosting in the Chinese medium set, whereas the present study shows both hedges and boosters elevated for the British condition. The model has not reproduced the finer hedge booster asymmetry reported in human writing studies. Instead, it appears to associate the British academic persona with a more generally marked and rhetorically managed register. The model's internal encoding is therefore partly concordant with the literature and partly a model specific recombination of discourse features.

The sentence level baseline is central to interpreting these results. No significant nationality differences appeared in the full generated surface text. The structural and stance contrasts identified through probing therefore cannot be treated as broad corpus wide asymmetries. They are concentrated at the token positions where the model's hidden states encode nationality most strongly. This is one of the clearest methodological results in the paper. Surface reading alone is insufficient for identifying the model's culturally differentiated encoding. The six cohort confound checks support the same conclusion. Medium effects within nationality were not significant, role did not predict phrase type or modifier structure, and the main nationality effects were therefore not reducible to medium or role. The CMI only result requires caution because it was partly driven by the small British CMI postdoc subgroup, but it does not overturn the overall pattern.

### *5.3 Domain vocabulary and culturally hollow writing*

The domain analysis provides the clearest link between the findings and the broader argument about culturally hollow writing. British distinctive positions clustered in evaluative and process oriented vocabulary, while Chinese distinctive positions clustered in sociocultural and internationalisation discourse. This asymmetry was robust in the five token window data and weaker in the smaller single token dataset. The pattern suggests that the model differentiates the two personas not only through isolated lexical items but through rhetorical register. British associated vocabulary is oriented toward evaluation, process, supervision, and the management of academic claims, whereas Chinese associated vocabulary is oriented more toward topic, content, and internationalisation.

The sentence level baseline makes the significance of this pattern much clearer. When the full generated surface text was analysed directly, none of the structural tests remained significant, none of the contrastive rhetoric hypotheses were confirmed, and none of the stance marker comparisons survived correction for multiple comparisons. Phrase type, UPOS, predicate type, hedging, boosting, modality, passive voice, and nationality by position interactions all returned null results at the sentence level. The model therefore shows statistically significant differentiation at probe selected positions, but not at the level of the full surface output. This contrast is what gives the argument about culturally hollow writing its force.

This pattern provides concrete evidence for what Section 2.2 describes as culturally hollow writing, in which output appears fluent and genre appropriate while remaining limited in cultural depth and situated meaning (Ahmad et al., 2024; Hyland, 2026). Ahmad et al. (2024) illustrate this problem in a different context by showing that ChatGPT can produce plausible responses about Hausa culture while failing to capture culturally grounded judgements. The present findings indicate that a similar issue can arise in academic discourse generation. The model has learned to associate the British condition with a default evaluative register and the Chinese condition with internationalisation rhetoric, but those associations do not amount to cultural knowledge in any deep sense. They are statistical regularities derived from training data. The results are therefore consistent with Hyland's (2026) argument that generative AI reproduces the formal markers of academic interaction without communicative intent, producing interactional fluency without relational grounding.

The domain findings also connect directly to the cultural alignment literature reviewed in Section 2.3. Tao et al. (2024) show that GPT model outputs align more closely with English speaking and Protestant European value profiles than with many other national contexts, and AlKhamissi et al. (2024) show that cultural alignment depends on prompt language, prompting, and pretraining composition. The present study complements these output level findings by identifying where cultural differentiation is encoded in the network and what linguistic features carry that signal. The layer by layer analysis indicates that lexical domain associations peak earlier in the network than the main structural nationality signal. The model therefore differentiates nationality first through lexical selection and later through more abstract syntactic and stylistic patterning. Vocabulary level interventions alone would not address deeper structural encoding.

### *5.4 Limitations*

Several limitations qualify the present findings. First, the probing methodology establishes recoverability rather than causality. As Belinkov (2022) emphasises, successful probing does not demonstrate that the model actively uses the encoded information during generation. No intervention method, such as activation patching or amnesic counterfactual probing, was applied here. A stronger causal claim would require direct manipulation of the identified representations and observation of the resulting effects on generated text.

Second, the corpus consists entirely of model generated text, so the representations being probed originate from the same system that produced the outputs. The study addresses this problem through triangulation against independent contrastive rhetoric research and against output level cultural alignment studies, but the validation remains indirect. Direct comparison with a purpose built corpus of human authored British and Chinese engineering research article introductions would provide a much stronger external benchmark for testing whether the token level patterns identified here correspond to measurable human writing differences.

Third, the probe architecture is deliberately simple. Only a linear logistic regression classifier with fixed regularisation was used, so the study cannot distinguish between information that is compactly linearly encoded and information that would only become recoverable through more complex decoding. This is the issue raised by Pimentel et al. (2020) and Voita and Titov (2020) in the probe complexity literature. Fourth, although the Stanza pipeline was audited and amended, some classification noise inevitably remains, especially for fine grained structural categories. Fifth, the weaker base to theory transfer result indicates that the nationality signal is not completely independent of prompt family and that more theoretically oriented prompts may activate somewhat different representational pathways.

### *5.5 Implications and future work*

The implications of the study are methodological, linguistic, and pedagogical. For interpretability research, the paper shows that probing can be extended beyond syntax and semantics to sociolinguistic and cultural representation. For applied linguistics, it provides mechanistic evidence that some culturally differentiated rhetorical tendencies are encoded internally in the model, not merely visible in outputs after decoding. For EAP and language pedagogy, it identifies a more direct educational risk. Students and teachers may encounter fluent, genre appropriate output that carries patterned cultural assumptions not detectable through ordinary surface evaluation.

The domain tested in the study is central to this point. EAP and pedagogical discourse is the site through which academic norms are transmitted to students through model texts, writing tasks, feedback, and assessment criteria. When multilingual engineering students use a language model in this domain, they are not simply receiving neutral language help. They may be receiving a statistically generated version of what the model has learned to associate with British or Chinese academic writing. This concern connects directly to the wider literature on algorithmic injustice and academic discourse. Bender et al. (2021) frame language models as sociotechnical systems shaped by data extraction and representational coverage. Noble (2018) and Benjamin (2019) show how apparently neutral systems can reproduce and entrench existing inequalities. Canagarajah (2002) and Lillis and Curry (2010) show that dominant Anglophone conventions can marginalise other rhetorical traditions. The present findings provide representational evidence for how such marginalisation may operate at the model level. British associated discourse clusters in evaluative and supervisory vocabulary, while Chinese associated discourse clusters in internationalisation rhetoric. This is not a neutral distribution. It reflects the structure of the English language training corpus, in which one academic profile is more readily available as a default evaluative register and the other is framed through more restricted thematic associations.

The implications for AI literacy follow directly. Long and Magerko (2020) define AI literacy as understanding, evaluating, and using AI responsibly. Pérez-Paredes et al. (2025) extend this to include ethics, language, ideology, and power. Tseng and Warschauer (2023) argue that students need to learn how to verify and integrate AI outputs critically, and Yan (2023) shows empirically that the value of AI generated feedback depends on how it is processed, discussed, and checked. The present findings give those recommendations a specific empirical basis. The nationality encoding identified here is not visible in the model's surface output, but is present in its hidden representations and is recoverable through the distributional properties of specific token positions at specific layers. A student or educator evaluating the model's output by reading it would not detect the encoded cultural assumptions. Critical AI literacy in this context therefore requires awareness that apparently neutral academic prose may still carry patterned cultural bias.

Several directions for further research follow naturally. Causal intervention methods should test whether the identified nationality linked representations actively shape generated text. The analysis should be extended to strictly technical engineering prompts to determine whether the same signal persists in more formulaic disciplinary prose. The present triangulation should be strengthened by comparison with a matched corpus of human authored British and Chinese engineering research article introductions. Larger models and models with different training data compositions should also be compared in order to determine whether the observed pattern is specific to Gemma 3 4b it or reflects a broader tendency in English language LLM training.

## 6. Conclusion

This study shows that Gemma 3 4b it encodes nationality-discriminative information in its hidden representations when generating persona-conditioned academic text. That encoding is structured, layer-specific, and linguistically characterisable. It is strongest in the middle to upper part of the network and is associated with systematic differences in modification, stance, and domain vocabulary at probe-selected token positions. At the same time, these differences do not remain statistically visible in the full generated surface text. The model therefore carries culturally differentiated rhetorical structure internally while producing output that appears broadly neutral at sentence level. This pattern is consistent with culturally hollow writing: interactionally fluent academic prose that suppresses the cultural assumptions involved in its generation. The findings extend probing methodology to a sociolinguistic attribute not previously examined in this way and show why the domain of EAP and language pedagogy matters. In educational settings, students and teachers may treat fluent model output as a neutral representation of academic practice even when patterned cultural encoding remains hidden beneath the surface.